\documentclass[10pt,twocolumn,letterpaper]{article}

\usepackage{cvpr}              
\usepackage{algorithm} 
\usepackage{algpseudocode}
\usepackage{comment}

\usepackage{graphicx}
\usepackage{pifont}
\usepackage{adjustbox}

\definecolor{darkgreen}{RGB}{1,50,32}
\definecolor{forestgreen}{RGB}{34,139,34}
\definecolor{cvprblue}{rgb}{0.21,0.49,0.74}
\usepackage[pagebackref,breaklinks,colorlinks,allcolors=cvprblue]{hyperref}

\def\paperID{32} 
\def\confName{CVPR}
\def\confYear{2025}

\title{AetherVision-Bench: An Open-Vocabulary RGB-Infrared Benchmark for Multi-Angle Segmentation across Aerial and Ground Perspectives }

\author{Aniruddh Sikdar*$\ ^{1}$, Aditya Gandhamal* $\ ^{2}$, Suresh Sundaram$^{1,3}$        
\\  
$^{1}$Robert Bosch Centre for Cyber Physical Systems, Indian Institute of Science, Bengaluru, India\\
$^{2}$Kotak IISc AI-ML Centre, Indian Institute of Science, Bengaluru, India\\
$^{3}$Department of Aerospace Engineering, Indian Institute of Science, Bengaluru, India\\
}

\begin{document}
\maketitle

\begin{abstract}
Open-vocabulary semantic segmentation (OVSS) involves assigning labels to each pixel in an image based on textual descriptions, leveraging world models like CLIP.
However, they encounter significant challenges in cross-domain generalization, hindering their practical efficacy in real-world applications. Embodied AI systems are transforming autonomous navigation for ground vehicles and drones by enhancing their perception abilities, and in this study, we present AetherVision-Bench, a benchmark for multi-angle segmentation across aerial, and ground perspectives, which facilitates an extensive evaluation of performance across different viewing angles and sensor modalities. We assess state-of-the-art OVSS models on the proposed benchmark and investigate the key factors that impact the performance of zero-shot transfer models. Our work pioneers the creation of a robustness benchmark, offering valuable insights and establishing a foundation for future research.
\end{abstract}

\section{Introduction}
\label{sec:intro}

Developing deep learning models with strong cross-domain generalization is essential for real-world computer vision tasks, including surveillance\cite{sikdar2023deepmao, sikdar2022fully}, disaster management \cite{john2024efficient}, and robotics across multiple domains \cite{john2025resource, sikdar2025saga, agrawal2025syn2real}. Open-vocabulary semantic segmentation (OVSS) emerges as a promising paradigm, enabling pixel-level classification from an open set of categories through the integration of natural language descriptions as semantic guidance.
Recent large-scale vision-language models, pre-trained on vast internet-scale data with natural language supervision, demonstrate strong generalization capabilities \cite{shi2024lca, radford2021learning, jia2021scaling, yuan2021florence}. Despite being trained on image-text pairs, these models rely on image-level supervision, limiting their pixel-level segmentation performance \cite{zhou2022extract, ding2022decoupling}.

\begin{figure}[t]
    \centering
    \includegraphics[scale=0.25]{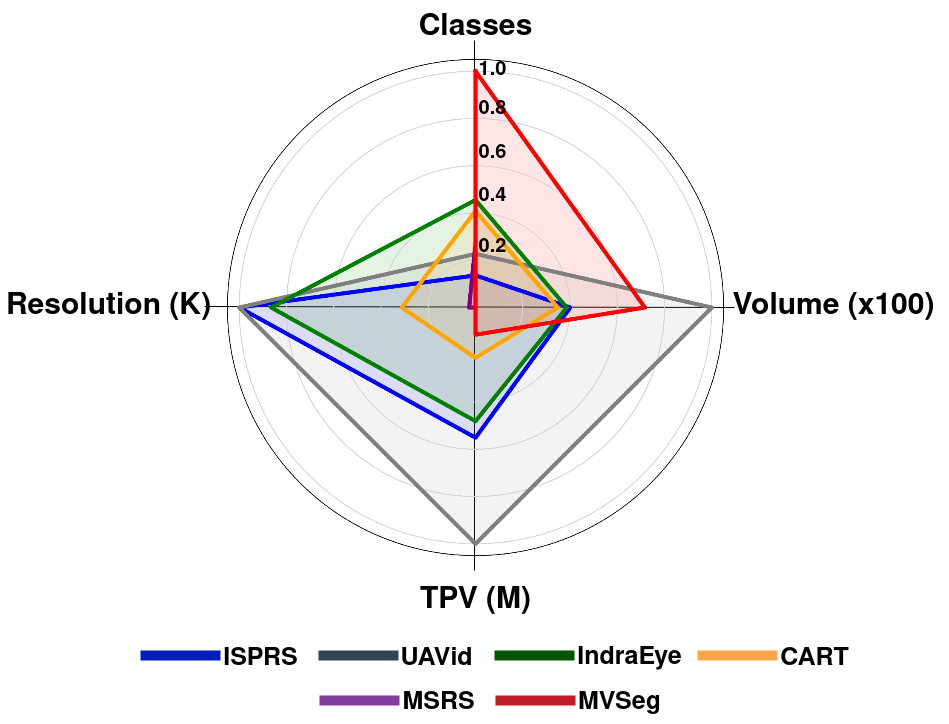}
    \setlength{\abovecaptionskip}{-5pt} 
    \setlength{\belowcaptionskip}{-11pt}
    \caption{Radar chart comparing test sets across AetherVision-Bench datasets, highlighting the number of classes, image resolution, Total Pixel Volume (TPV), and overall dataset volume. While the maximum number of classes is 26, the data is normalized based on resolution for visualization purposes. The chart illustrates the balanced nature of the benchmark across these key dimensions.}
    \label{fig:radar_plot}
\end{figure}

Recent studies have improved these models for segmentation tasks through innovative approaches, including (1) mask proposal networks  \cite{ghiasi2022scaling, xu2022simple, xu2023side, yu2023convolutions} and (2) using cost aggregation methods \cite{cho2024cat}.  \cite{blumenstiel2024mess} observes that pixel-level classification performs poorly on unseen classes, particularly in the presence of significant domain shifts. This limitation becomes increasingly pronounced as the number of unseen categories grows \cite{ding2022decoupling}, with open-vocabulary models exhibiting heightened vulnerability to overfitting during downstream task adaptation \cite{wortsman2022robust,kumar2022fine}. Although OVSS models effectively capture open-set visual concepts for segmentation, their zero-shot generalization performance degrades notably when faced with major category shifts, distribution, or domain shifts between large-scale pretraining data and downstream tasks \cite{zhang2024dept}. This constrains the practical deployment of open-vocabulary models in real-world scenarios.

\begin{figure}[t]
    \centering
    \includegraphics[scale=0.25]{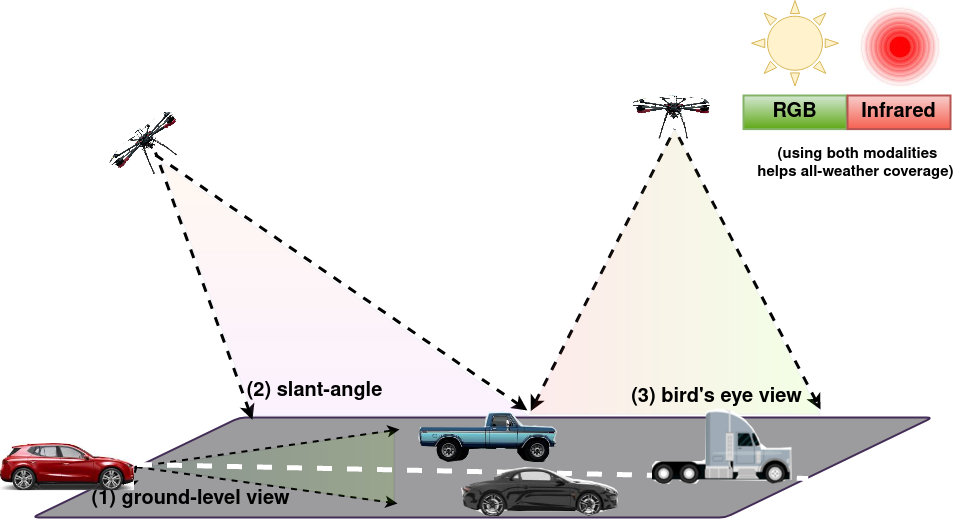}
    \caption{Comparison of viewpoints across three settings: (1) ground-level, (2) slant-angle, and (3) bird’s-eye view. We evaluate methods across these perspectives while accounting for the multi-modal nature of the datasets to ensure robustness under adverse weather conditions.}
    \label{fig:look-angles}
\end{figure}

Embodied AI systems are revolutionizing autonomous navigation for ground vehicles \cite{udupa2024mrfp} and drones \cite{sikdar2025saga} by enabling advanced perception capabilities. These systems should be designed to harness both visiable (RGB) and infrared (IR) sensor data, ensuring resilient performance in adverse weather conditions and low-light environments \cite{sikdar2024skd}. They should be robust to sensor modality changes, such as transitions from RGB to IR \cite{sikdar2024ssl}, ensuring reliable performance under diverse conditions including day-night cycles, fog, rain, and glare \cite{sikdar2025ogp}. 
However, existing standard benchmarks primarily evaluate in-domain tasks, offering limited perspective on cross-domain performance.This limits understanding of zero-shot segmentation models' generalization to novel domains and their suitability for complex datasets with varied sensors or vocabularies. Existing benchmarks, such as the Multi-domain Evaluation of Semantic Segmentation (MESS) \cite{blumenstiel2024mess}, have been developed to assess model performance across a broader range of complex, domain-specific datasets. However, leveraging foundation models like CLIP \cite{radford2021learning} for advanced perception presents three key challenges that remain unexplored in previous benchmarks: (1) fine-tuning on limited training dataset, for downstream segmentation tasks while preserving pretrained knowledge, (2) handling significant domain shifts arising from out-of-distribution data, and (3) accommodating substantial variations between seen and unseen classes, particularly those arising from differing viewing perspectives. These challenges are integral to ensuring the robustness  of models across diverse downstream tasks.

To address this, we introduce AetherVision-Bench: an open-vocabulary RGB-Infrared benchmark for multi-angle segmentation across drones, satellites, and ground perspectives, as shown in Figure \ref{fig:look-angles}. We evaluate state-of-the-art OVSS models across six RGB and four IR datasets, carefully chosen to encompass a diverse set of classes and out-of-distribution domains. These datasets are strategically selected to include various object viewpoints, ensuring a comprehensive assessment of model performance under different conditions. We evaluate these models under diverse training regimes, including COCO-Stuff, which offers limited object viewpoint diversity, and RGB datasets featuring both eye-level and slant-angle perspectives, validated comprehensively on the AetherVision benchmark. This underscores the critical need for robust models capable of handling diverse illumination conditions, sensor shifts, noisy sensor inputs, and varying viewing angles.  The radar plot in Fig. \ref{fig:radar_plot} illustrates the balanced distribution of datasets included in AetherVision-Bench. Our benchmark aims to advance zero-shot segmentation and boost real-world OVSS performance.

\begin{table}[]
\centering
\begin{tabular}{l|cc|ccc}
\hline
                                  & {\textbf{Modality}} & {\textbf{Viewpoint}} \\ \cline{2-6} 

\multicolumn{1}{c|}{{\textbf{Dataset}}} & \textbf{RGB}                   & \textbf{IR}                   & \textbf{BEV}         & \textbf{SA}        & \textbf{GL}        \\ \hline
ISPRS                                                                           & \ding{51}                              & \ding{55}                             & \ding{51}                    & \ding{55}                  & \ding{55}                  \\
UAVid                                                                           & \ding{51}                              & \ding{55}                             & \ding{55}                    & \ding{51}                  & \ding{55}                  \\
IndraEye                                                                        & \ding{51}                              & \ding{51}                             & \ding{55}                    & \ding{51}                  & \ding{55}                  \\
CART                                                                            & \ding{51}                              & \ding{51}                             & \ding{55}                    & \ding{51}                  & \ding{55}                  \\
MSRS                                                                            & \ding{51}                              & \ding{51}                             & \ding{55}                    & \ding{55}                  & \ding{51}                  \\
MVSeg                                                                           & \ding{51}                              & \ding{51}                             & \ding{55}                    & \ding{55}                  & \ding{51}                  \\ \hline
\end{tabular}
\caption{Comparison of datasets with respect to modalities and different viewpoints viz. birds-eye-view (BEV), slant-angle (SA) and ground-level (GL).}
\label{table:viewpoint}
\end{table}

Our key contributions are summarized as follows:  (1) We develop a taxonomy grounded in quantitative analysis of diverse semantic segmentation datasets and introduce a novel benchmark for out-of-domain OVSS. This benchmark is designed to rigorously assess advanced perception systems across a wide range of scenarios, including ground-level driving scenes, aerial views, and bird’s-eye perspectives. (2) We present a comprehensive evaluation of state-of-the-art zero-shot models on AetherVision-Bench, revealing that OVSS performance is heavily influenced by class semantics, textual similarity, and sensor modality.

\begin{table*}[]
\resizebox{1.03\textwidth}{!}{%
\begin{tabular}{c|ccccccc|ccccc}
\hline

                                 & \multicolumn{7}{c|}{\textbf{RGB}}                                                                                                                                   & \multicolumn{5}{c}{\textbf{IR}}                                                                                                   \\ \cline{2-13} 
 
{\textbf{Method}} & \textbf{ISPRS} & \textbf{UAVid} & \textbf{IndraEye} & \textbf{CART}  & \textbf{MSRS}  & \multicolumn{1}{c|}{\textbf{MVSeg}} & \textbf{mIoU} & \textbf{IndraEye} & \textbf{CART}  & \textbf{MSRS}  & \multicolumn{1}{c|}{\textbf{MVSeg}} & \textbf{mIoU} \\ \hline
SAN                                                       & 37.56          & \underline{41.65}    & 7.84              & 37.36          & 20.35          & \multicolumn{1}{c|}{21.96}                                  & 27.79                                 & 4.91              & \underline{25.35}    & \textbf{19.93} & \multicolumn{1}{c|}{11.93}                                  & \underline{15.53}                           \\
Zegformer                                                 & 14.04          & 17.85          & 12.92             & 17.18          & 12.48          & \multicolumn{1}{c|}{11.6}                                   & 14.34                                 & 3.21              & 3.34           & 7.38           & \multicolumn{1}{c|}{3.4}                                    & 4.33                                  \\
OVSeg                                                     & 17.52          & 33.69          & 13.4              & 31.67          & 19.36          & \multicolumn{1}{c|}{14.87}                                  & 21.75                                 & 6.82              & 16.46          & 13.91          & \multicolumn{1}{c|}{5.29}                                   & 10.62                                 \\
CAT-Seg                                                   & \underline{45.36}    & 41.55          & 12.84             & \underline{39.66}    & \textbf{26.63} & \multicolumn{1}{c|}{\underline{22.27}}                            & \underline{31.39}                           & 6.7               & 20.02          & 17.85          & \multicolumn{1}{c|}{\underline{12.62}}                            & 14.30                                 \\ \hline
OVSeg-L                                                   & 31.03          & 38.79          & \textbf{16.44}    & 35.21          & 19.98          & \multicolumn{1}{c|}{19.58}                                  & 26.84                                 & \textbf{12.48}    & 20.7           & 15.53          & \multicolumn{1}{c|}{9.01}                                   & 14.43                                 \\
CAT-Seg-L                                                 & \textbf{50.92} & \textbf{41.9}  & \underline{16.22}       & \textbf{39.85} & \underline{20.77}    & \multicolumn{1}{c|}{\textbf{26.47}}                         & \textbf{32.69}                        & \underline{11.74}       & \textbf{25.84} & \underline{19.44}    & \multicolumn{1}{c|}{\textbf{15.32}}                         & \textbf{18.09}                        \\ \hline
\end{tabular}
}
\caption{Performance comparison of state-of-the-art OVSS methods trained on COCO-Stuff and evaluated on both RGB and IR modalities of the AetherVision-Bench.}
\label{table:mscoco}
\end{table*}

\section{Related Works}

\textbf{Open-vocabulary semantic segmentation (OVSS)} Large-scale pre-training has revolutionized computer vision by training models on image-text pairs, enabling representation of visual and textual semantics in a shared space. Open-vocabulary semantic segmentation (OVSS) aims to bridge the gap between image-level and pixel-level understanding by enabling segmentation of unseen classes. OVSS models are generally categorized into two main types: (i) mask-proposal networks \cite{ghiasi2022scaling, ding2022open, xu2023open, xu2022simple, xu2023side, yu2023convolutions}, and (ii) cost-aggregation networks \cite{cho2024cat}. Mask-proposal approaches generate class-agnostic region proposals, which are then semantically aligned with CLIP-derived textual embeddings to facilitate open-vocabulary classification at the pixel level. For instance, OpenSeg \cite{ghiasi2022scaling} leverages local image regions to form proposals, whereas models like ZegFormer \cite{ding2022decoupling} and ZSseg \cite{xu2022simple} adopt a two-stage architecture to address this task.   CAT-Seg \cite{cho2024cat} enhances robustness by using similarity scores to combine CLIP’s image and text embeddings into a cost volume, rather than relying on direct embeddings. OVSS models are commonly evaluated on standard datasets like ADE20K \cite{zhou2019semantic} and Pascal VOC \cite{everingham2015pascal}, which primarily feature everyday, in-domain images. However, such evaluations fail to capture real-world conditions with inherent domain shifts and environmental diversity.  \\
\textbf{Related Benchmarks} OVSS models are typically evaluated on datasets composed exclusively of in-domain images \cite{zhou2019semantic,everingham2015pascal}, which are considered the de facto standard for model evaluation \cite{ghiasi2022scaling, liang2023open}. Few works have explored broader dataset diversity. The Segmentation in the Wild (SegInW) benchmark \cite{zou2023generalized} includes 25 datasets, however, most of them still feature everyday, in-domain images, with only two notable exceptions. Inspired by the HELM benchmark \cite{liang2022holistic} for evaluating large language models, the MESS benchmark \cite{blumenstiel2024mess} was subsequently introduced for assessing OVSS models. MESS motivates a comprehensive evaluation across a wide spectrum of domain-specific datasets, including medicine, engineering, earth observation, biology, and agriculture.. However, a key limitation lies in the relatively low number of classes within these datasets, resulting in high random mIoU scores. Additionally, the benchmark empirically demonstrates that current OVSS models struggle to segment objects in infrared imagery, highlighting sensor-shift as a critical challenge. Most state-of-the-art models exhibit suboptimal performance and offer limited support for generalization in open-world settings. To the best of our knowledge, open-vocabulary semantic segmentation (OVSS) models have not been systematically evaluated on datasets that account for sensor shifts and variations in look-angle, as commonly encountered in drones and autonomous vehicles. This work presents the first comprehensive effort to address these challenges, aiming to provide new insights and advance the robustness of OVSS models in practical settings.

\begin{table*}[]
\resizebox{1.02\textwidth}{!}{%
\begin{tabular}{c|ccccccc|ccccc}
\hline
 
                                  & \multicolumn{7}{c|}{\textbf{RGB}}                                                                                                                                   & \multicolumn{5}{c}{\textbf{IR}}                                                                                                   \\ \cline{2-13} 

{\textbf{Method}} & \textbf{ISPRS} & \textbf{UAVid} & \textbf{Indraeye} & \textbf{CART}  & \textbf{MSRS}  & \multicolumn{1}{c|}{\textbf{MVSeg}} & \textbf{mIoU} & \textbf{IndraEye} & \textbf{CART}  & \textbf{MSRS}  & \multicolumn{1}{c|}{\textbf{MVSeg}} & \textbf{mIoU} \\ \hline
SAN                                                       & 10.11          & \underline{13.86}    & 5.02              & \underline{12.27}    & 68.63          & \multicolumn{1}{c|}{\underline{7.77}}                             & 19.61                                 & 2.29              & \underline{5.9}      & 30.18          & \multicolumn{1}{c|}{\underline{2.2}}                              & 10.14                                 \\
Zegformer                                                 & 4.59           & 4.76           & \underline{6.68}        & 3.41           & \underline{71.49}    & \multicolumn{1}{c|}{4.9}                                    & 15.97                                 & \textbf{8.06}     & 3.76           & \textbf{34.83} & \multicolumn{1}{c|}{1.86}                                   & \underline{12.13}                           \\
OVSeg                                                     & \underline{10.45}    & 4.18           & 5.16              & 2.81           & 65.0           & \multicolumn{1}{c|}{4.55}                                   & \underline{21.29}                           & 3.81              & 2.65           & 28.66          & \multicolumn{1}{c|}{1.39}                                   & 11.29                                 \\
CAT-Seg                                                   & \textbf{18.34} & \textbf{28.34} & \textbf{7.06}     & \textbf{22.87} & \textbf{71.93} & \multicolumn{1}{c|}{\textbf{9.24}}                          & \textbf{26.30}                        & \underline{4.33}        & \textbf{10.34} & \underline{33.59}    & \multicolumn{1}{c|}{\textbf{4.47}}                          & \textbf{13.18}                        \\ \hline
\end{tabular}
}
\caption{Performance comparison of state-of-the-art OVSS methods trained on MSRS (RGB) and evaluated on both RGB and IR modalities of the AetherVision-Bench.}
\label{table:MSRS}
\end{table*}

\begin{table*}[]
\resizebox{1.01\textwidth}{!}{%
\begin{tabular}{c|ccccccc|ccccc}
\hline

                                 & \multicolumn{7}{c|}{\textbf{RGB}}                                                                                                                                 & \multicolumn{5}{c}{\textbf{IR}}                                                                                                  \\ \cline{2-13} 
{\textbf{Method}} & \textbf{ISPRS} & \textbf{UAVid} & \textbf{Indraeye} & \textbf{CART} & \textbf{MSRS} & \multicolumn{1}{c|}{\textbf{MVSeg}} & \textbf{mIoU} & \textbf{IndraEye} & \textbf{CART} & \textbf{MSRS}  & \multicolumn{1}{c|}{\textbf{MVSeg}} & \textbf{mIoU} \\ \hline
SAN                                                       & \underline{13.33}    & \underline{15.57}    & \textbf{75.84}    & 6.04          & \textbf{9.28} & \multicolumn{1}{c|}{\underline{6.12}}                             & \textbf{21.03}                        & 23.24             & 3.21          & \textbf{10.51} & \multicolumn{1}{c|}{\underline{3.12}}                             & \underline{10.02}                           \\
Zegformer                                                 & 7.80           & 7.32           & \underline{75.52}       & 3.51          & 4.72          & \multicolumn{1}{c|}{2.9}                                    & 16.96                                 & \underline{28.05}       & 2.31          & 2.66           & \multicolumn{1}{c|}{0.48}                                   & 8.38                                  \\
OVSeg                                                     & 12.3           & 16.23          & 61.8              & \underline{9.0}       & \underline{5.68}    & \multicolumn{1}{c|}{4.12}                                   & 18.19                                 & 22.41             & \underline{5.12}    & 2.53           & \multicolumn{1}{c|}{1.55}                                   & 7.90                                  \\
CAT-Seg                                                   & \textbf{17.2}  & \textbf{17.62} & 58.88             & \textbf{15.3} & 4.03          & \multicolumn{1}{c|}{\textbf{8.84}}                          & \underline{20.31}                           & \textbf{33.90}    & \textbf{8.07} & \underline{4.99}     & \multicolumn{1}{c|}{\textbf{4.94}}                          & \textbf{12.97}                        \\ \hline
\end{tabular}
}
\caption{Performance comparison of state-of-the-art OVSS methods trained on IndraEye (RGB) and evaluated on both RGB and IR modalities of the AetherVision-Bench.}
\label{table:Indraeye}
\end{table*}

\section{AetherVision-Benchmark}

\textbf{Problem Formulation} Let $I$ represent an image associated with a set of  classes  $C$ = \{$C_{1}$, $C_{2}$, ..., $C_{N}$\}, where each class $C_{i}$ is defined by a textual label. OVSS models aim to assign a class label $C_{i}$ for each pixel in an image $I$, where the candidate class set $N$ can vary at inference and may include novel categories unseen during fine-tuning on the training data.

\subsection{AetherVision-Benchmark}
Real-world OVSS deployment faces key robustness challenges often overlooked in current research. These challenges highlight the need to improve OVSS model robustness for real-world use. To address this, we propose AetherVision-Bench: An Open-Vocabulary RGB-Infrared Benchmark for Multi-Angle Segmentation Across Drones, Satellites, and Ground Perspectives.

We filter the available datasets based on two exclusion criteria: (1) viewing angle and (2) sensor modality. Guided by the proposed taxonomy, we categorized datasets into three primary look-angle clusters: (1) \textit{ground-level}, (2) \textit{slant-angle}, and (3) \textit{bird's-eye} view, as shown in Table \ref{table:viewpoint}. This facilitates the assessment of a model’s ability to segment objects under significantly varying viewpoints. Subsequently, we evaluate on both RGB and IR modalities to promote robustness under varying environmental and weather conditions. For each viewpoint category, we evaluate the models’ segmentation performance under diverse domain and sensor shifts. For the bird’s-eye view, we selected the ISPRS \cite{swissphoto2012isprs} dataset. To capture varying slant-angle perspectives, we included UAVid \cite{lyu2020uavid}, IndraEye \cite{sikdar2025saga}, and CART \cite{lee2024caltech} datasets, where both IndraEye and CART also provide corresponding IR imagery. For ground-level viewpoints, we use the MSRS \cite{tang2022piafusion} and MVSeg \cite{ji2023multispectral} datasets, with their IR modalities. We adopt three evaluation settings: (1) training OVSS models on COCO-Stuff \cite{caesar2018coco}, (2) training on the slant-angle RGB IndraEye dataset, and (3) training on the ground-level MSRS dataset. These settings assess the models’ capability to generalize across diverse look-angles by training on a specific viewpoint and validating on all three. Notably, settings (2) and (3) introduce an additional challenge of limited training samples. We selected these datasets based on official test splits, high-quality annotations, and sufficient image resolution. No ethical concerns were raised, and the datasets were used in compliance with their respective licenses. This benchmark is ideal for evaluating model robustness across diverse look angles, with challenges like weather changes, sensor degradation, and complex environments.

\section{Experiments}

\textbf{Implementation} All models are evaluated using the mean intersection over union (mIoU), using their official implementations, and pretrained weights. We use the AetherVision-Bench to evaluate state-of-the-art OVSS models, including OVSeg \cite{liang2023open}, Zegformer \cite{ding2022decoupling}, SAN \cite{xu2023side}, and CAT-Seg \cite{cho2024cat}, which feature various architectural approaches (e.g., two-stage mask-based, one-stage mask-based, and pixel-based).\\
\subsection{Multi-domain OVSS} Table \ref{table:mscoco} presents a comparison of OVSS models on the AetherVision-Bench, trained on the COCO-Stuff dataset. Table \ref{table:MSRS} compares the models when trained on the MSRS dataset's training set, using the RGB modality. As shown in Table \ref{table:Indraeye}, training on the slant-angle RGB samples from the IndraEye dataset leads to a significant drop in performance.\\
\textbf{In-domain vs multi-domain} Most OVSS models are typically evaluated on in-domain datasets. However, AetherVision-Bench's real-world benchmarking is particularly effective for assessing the robustness under conditions that closely resemble those found in real-world deployments. As noted, the CAT-Seg model performs better than other models, when trained on COCO-Stuff and MSRS datasets, but its performance significantly drops when trained on the slant-angle view in the benchmark.\\
\textbf{Sensor Domain Shift} As seen in all the tables across all settings, the robustness of all models to sensor shifts is poor. The overall performance decreases by nearly half, as demonstrated in Table \ref{table:mscoco}, \ref{table:MSRS}, and \ref{table:Indraeye}.
Even when models are trained on the MSRS RGB dataset, their performance on co-registered IR pairs drops to less than half. For noisy sensors, as seen with the MVSeg dataset, performance deteriorates even further.\\
\textbf{Evaluation under diverse viewing angles} Models trained on COCO-Stuff or ground-level datasets demonstrate limited effectiveness on drone-view imagery. Similarly, those trained on slant-angle data such as the IndraEye dataset show poor generalization to ground-level datasets like MSRS and MVSeg. This indicates that OVSS models must depend heavily on their pre-training to generalize across different viewpoints, incase these viewing points are not available during training for downstream tasks.

\section{Conclusion} 
This paper presents a comprehensive benchmark for evaluating the robustness of OVSS models, essential for their real-world deployment. We introduce the AetherVision-Bench, a Benchmark for multi-angle segmentation across various perspectives. This benchmark evaluates models across two challenging scenarios: viewing angle and sensor modality.  It is designed to address real-world complexities, making it ideal for embodied AI systems that enhance perception and revolutionize autonomous navigation for ground vehicles and drones.
{
    \small
    \bibliographystyle{ieeenat_fullname}
    \bibliography{main}
}


\end{document}